# Exploring a Fine-Grained Multiscale Method for Cross-Modal Remote Sensing Image Retrieval

Zhiqiang Yuan, *Student Member, IEEE*, Wenkai Zhang, *Member, IEEE*, Kun Fu, *Member, IEEE*, Xuan Li, Chubo Deng, Hongqi Wang, *Member, IEEE*, and Xian Sun, *Senior Member, IEEE*

*Abstract*—Remote sensing (RS) cross-modal text–image retrieval has attracted extensive attention for its advantages of flexible input and efficient query. However, traditional methods ignore the characteristics of multiscale and redundant targets in RS image, leading to the degradation of retrieval accuracy. To cope with the problem of multiscale scarcity and target redundancy in RS multimodal retrieval task, we come up with a novel asymmetric multimodal feature matching network (AMFMN). Our model adapts to multiscale feature inputs, favors multisource retrieval methods, and can dynamically filter redundant features. AMFMN employs the multiscale visual self-attention (MVSA) module to extract the salient features of RS image and utilizes visual features to guide the text representation. Furthermore, to alleviate the positive samples ambiguity caused by the strong intraclass similarity in RS image, we propose a triplet loss function with dynamic variable margin based on prior similarity of sample pairs. Finally, unlike the traditional RS image-text dataset with coarse text and higher intraclass similarity, we construct a fine-grained and more challenging Remote sensing Image-Text Match dataset (RSITMD), which supports RS image retrieval through keywords and sentence separately and jointly. Experiments on four RS text–image datasets demonstrate that the proposed model can achieve state-of-the-art performance in cross-modal RS text–image retrieval task.

## I. Introduction

IN RECENT years, remote sensing (RS) images have qualitatively and quantitatively improved our perception of the earth [1], [2]. Even though the vast RS data contain abundant information, how to choose valuable knowledge for human beings is a very challenging task. Automatically, designing an effective RS image retrieval method to perform retrieval task has become the focus of more and more researchers [3], [4].

Generally, the RS image retrieval methods can be divided into two types: the unimodal retrieval and the multimodal retrieval [5]. In RS unimodal retrieval, the query data and RS data belong to the same modality. Chen *et al.* [6] utilized a hash algorithm to establish a new retrieval method, which improves the retrieval performance in efficiency. Demir and Bruzzone [7] analyzed the validity of hash retrieval from accuracy and time of retrieval and proposed two kernel-based nonlinear hash methods. RS multimodal retrieval means that query data and RS data are in different modalities. For example, Guo *et al.* [8] designed an image and audio similarity calculation model, which is based on the pretrained convolution neural network (CNN) and AudioNet and applies a neural network for fusion and classification. Compared with RS unimodal retrieval, the RS multimodal retrieval is more challenging since it requires to map different modal data into the uniform measurable space. In recent years, RS multimodal retrieval has become one of the research hotspots.

Cross-modal RS text–image retrieval plays an important role in RS multimodal retrieval. In the last decades, manual captions were usually used to provide tags for each RS image, and then, the query text was matched with the tagged captions [9]. With the rapid growth of RS images, manual captions have become increasingly time-consuming, and researchers have paid more attention to automatic image captions [10], [11]. For example, Shi and Zou [12] made use of the fully convolution network to construct an RS image caption framework. Although the retrieval method of generating caption solves human resources annotation, it may still leave some retrieval shortcomings. On the one hand, two-stage retrieval mode makes it difficult to avoid the loss of abundant information in the middle stage [13]. On the other hand, coarse caption generated by machine may not be used as a good representation of the RS image [14]. Consequently, the traditional RS text–image retrieval method may not be the optimal choice in cross-modal RS image retrieval task.

In recent years, some retrieval methods [15]–[17] based on deep learning have been proposed, and they calculate the similarity between images and texts directly. However, there are still three challenges when these methods are applied to cross-modal RS text–image retrieval completely. First, RS image with numerous objects often contains a large amount of background content unrelated to the description subject.

Manuscript received March 20, 2021; revised April 15, 2021; accepted May 5, 2021. This work was supported by the National Science Fund for Distinguished Young Scholars under Grant 67125105. *(Corresponding author: Xian Sun.)*

Zhiqiang Yuan and Xuan Li are with the Aerospace Information Research Institute, Chinese Academy of Sciences, Beijing 100190, China, also with the Key Laboratory of Network Information System Technology (NIST), Institute of Electronics, Chinese Academy of Sciences, Beijing 100190, China, also with the University of Chinese Academy of Sciences, Beijing 100190, China, and also with the School of Electronic, Electrical and Communication Engineering, University of Chinese Academy of Sciences, Beijing 100190, China (e-mail: yuanzhiqiang19@mails.ucas.ac.cn; lixuan173@mails.ucas.ac.cn).

Wenkai Zhang, Kun Fu, Chubo Deng, Hongqi Wang, and Xian Sun are with the Aerospace Information Research Institute, Chinese Academy of Sciences, Beijing 100190, China, and also with the Key Laboratory of Network Information System Technology (NIST), Institute of Electronics, Chinese Academy of Sciences, Beijing 100190, China (e-mail: zhangwk@aircas.ac.cn; kunfuiecas@gmail.com; dengcb@aircas.ac.cn; wiecas@sina.com; sunxian@mail.ie.ac.cn).

Digital Object Identifier 10.1109/TGRS.2021.3078451

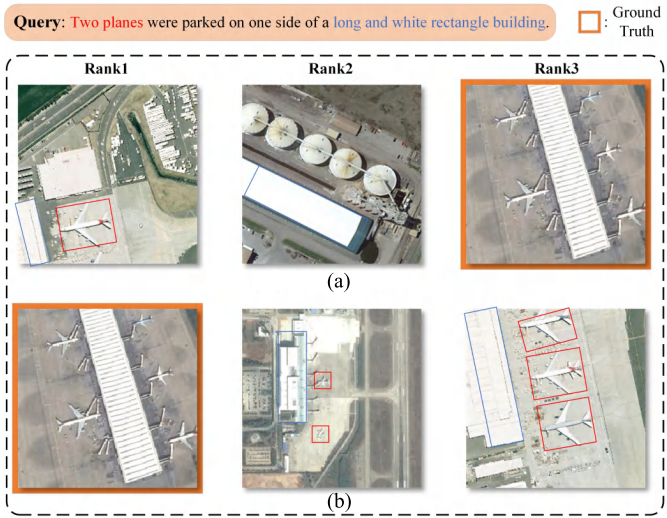

Fig. 1. Comparison of traditional and our retrieval results. Compared with the single-scale images retrieved by traditional method, we can obtain multiscale retrieval results as Rank2 of (b). Simultaneously, when the retrieval result is wrong, the error result obtained by the traditional method is quite different from the ground truth. However, the proposed triplet loss makes the error result similar to the ground truth as much as possible to ensure the robustness of the retrieval model.

However, natural scene images tend to have salient objects compared with RS images. Text–image retrieval methods in the natural domain ignore the filter of redundant features, which hinders the model to understand the content of RS image. Therefore, how to obtain salient features in the RS image has become an urgent problem. At the same time, the methods in natural scenes fail to fully consider the multiscale information of RS [18], resulting in insufficient utilization of multiscale features and unsatisfactory retrieval accuracy. Due to the multiscale and target redundancy in RS image, we attempt to extract salient features using a network with multiscale and dynamic filtering functions. To get better text features, we first use the filtered image features to guide the text representation dynamically and then apply this method to different retrieval tasks to obtain more flexible input.

Second, different from natural scenes, RS scenes have a strong intraclass similarity. A text may be corresponding to multiple negative sample images that are very similar to the ground truth. We call such negative samples soft positive samples. Due to the existence of soft positive samples, the model will be ambiguous about the optimization objective during training. We define this problem as the positive sample ambiguity. To alleviate the abovementioned problem, we transform the hard margin in the traditional method into the soft margin based on the prior similarity between sample pairs, which can adaptively change the fixed margin in the traditional triplet loss function and further improve the retrieval performance.

Third, the text in traditional RS scenes datasets is usually coarsely contracted with natural scenes, which makes traditional datasets unsuitable for text–image retrieval tasks because of their high intraclass similarity. To solve this issue, we construct a fine-grained and more challenging dataset to minimize the intraclass similarity. Simultaneously, we add keywords attribute to the dataset to enhance generalization ability for multiple retrieval tasks.

The main contributions of our work are as follows:

1) In order to solve the problem of multiscale scarcity and target redundancy in RS multimodal retrieval tasks, we design an asymmetric multimodal feature matching network (AMFMN). The proposed cross-modal method adapts to multiscale feature inputs, supports multisource retrieval methods, and can dynamically filter redundant features. AMFMN utilizes the multiscale visual self-attention (MVSA) module to extract the salient features of RS image and uses the visual features to guide the text representation, which achieves competitive results on several RS image-text datasets.

2) Aiming to alleviate the positive sample ambiguity caused by the strong intraclass similarity in RS image, we design a triplet loss function with a dynamic variable margin based on the sample pairs' prior similarity. The experiment results verify the feasibility of our work.

3) Compared with the traditional RS image-text dataset with coarse text and higher intraclass similarity, we construct a fine-grained and more challenging Remote Sensing Image-Text Match dataset (RSITMD). RSITMD has more scene changes and higher fine-grained caption. Besides, the new keywords attribute can be further applied to the task of RS text retrieval by keywords.

We have conducted plentiful experiments to compare with popular algorithms and explore the reasons why AMFMN can achieve state-of-the-art performance, as shown in Fig. 1. We also use the AMFMN to carry out location experiments, and the results show the effectiveness of our model.

The rest of this article is organized as follows. In Section II, we briefly summarize some previous works in the field of cross-modal RS image retrieval. In Section III, we introduce the implementation details of the AMFMN model and the dynamic variable triplet loss. Furthermore, we propose the fine-grained and more challenging RSITMD dataset in Section IV. In Section V, we give a quantitative comparison between the proposed model and other baseline models, and experiments results have demonstrated the superiority of our model in RS text–image retrieval. Finally, the conclusion is given in Section VI. The code of the AMFMN and proposed RSITMD dataset will be open to access soon.[1]

## II. RELATED WORK

In this section, we review the previous research on cross-modal retrieval of RS images. We will summarize from three aspects: image–image retrieval, audio–image retrieval, and text–image retrieval.

### A. Image–Image Cross-Modal RS Image Retrieval

Image–image cross-modal RS image retrieval refers to retrieve RS image by using RS image of other modalities, such as using synthetic aperture radar images to retrieve RS image in natural scenes. Li *et al*. [19] designed a new

---
[1]https://github.com/xiaoyuan1996/AMFMN

CNN based on source-invariant hash method and retrieved cross-modal RS image on dual-source RS image datasets. Zhang *et al.* [20] and Zhou *et al.* [21] exploited using image content and low-level features to retrieve RS image. Xiong *et al.* [22] increased the diversity of training methods by converting three channels of optical images into four different types of single-channel images and combined triplet loss with hash functions to improve the retrieval efficiency. Demir and Bruzzone [7] introduced hashing-based approximate nearest neighbor search to retrieve RS images fastly and accurately. Xiong *et al.* [23] proposed a cycle identity generation adversarial network to reduce data drift during multisource RS image retrieval. Even though the image–image cross-modal RS image retrieval task is quite mature, it is limited by the low-level features acquisition [24]. Therefore, the image–image retrieval method may not be able to perform retrieval efficiently in specific retrieval scenarios even if it can obtain better retrieval results.

### B. Audio–Image Cross-Modal RS Image Retrieval

Considering the audio for RS image retrieval, Mao *et al.* [25] produced a large-scale RS image dataset, which contains numerous hand-annotated voice captions, and performed cross-modal RS audio–image retrieval. Chen and Lu [26] proposed a deep triplet-based hashing to integrate hash code learning and relative relationship learning into the end-to-end network. Furthermore, Guo *et al.* [8] designed an audio–image similarity calculation model and then used a neural network for fusion and classification. Chen *et al.* [27] solved the problem of insufficient utilization of cross-modal similarity learning data by using the multiscale context information. Although the audio–image retrieval method brings excellent convenience to users, a certain amount of data noise may be generated due to the direct inputs of audio signal.

### C. Text–Image Cross-Modal RS Image Retrieval

Early RS text–image retrieval methods manually tag each RS image at first and then calculate the text similarity between the query text and the labeled text in the retrieval process in order to rank the results. With the increase of RS images, it takes much time and energy to manually label them. Meanwhile, with the development of automatic image caption, caption-based retrieval methods have become the mainstream of RS image retrieval [28]–[30]. Shi and Zou [12] designed an RS image caption framework by modeling the interaction of RS image attributes. Lu *et al.* [31] proposed the largest dataset RSICD in RS image caption. Hoxha *et al.* [32] used the CNN-RNN framework that combined with beam search to generate multiple captions for the target image and then selected the best caption by utilizing prior similarity. Li *et al.* [11] proposed a truncation cross entropy (TCE) loss to alleviate the overfitting problem and discussed the phenomenon of overfitting in RSICD. Lu *et al.* [33] proposed a sound active attention framework for more specific caption generation. Even though caption-based RS image retrieval is relatively mature, the process still has some drawbacks. Since the generated sentence is relatively coarse, the inevitable information loss in the caption generation stage and text similarity matching stage reduces the retrieval accuracy.

Recently, some methods are proposed to calculate a cross-modal similarity directly in natural scenes. After encoding the images and text, Faghri *et al.* [34] used a triplet loss function to minimize the distance between similar images and text. Lee *et al.* [35] tried to align the region in the image with the word in the caption to calculate the similarity. Wang *et al.* [36] presented a fusion model that is based on rank decomposition to calculate the similarity between images and text. Even if the methods of calculating the similarity between images and text directly are mature in the natural field, such methods are rare in RS scenes. Abdullah *et al.* [37] proposed a deep bidirectional triplet network for the RS text–image embedded similarity calculations. As far as we know, this is the only method for cross-modal text–image retrieval in the field of RS.

## III. METHOD

This article studies the text–image cross-modal retrieval in RS. Fig. 2 shows the complete architecture of the AMFMN. We will introduce AMFMN from five parts: 1) formulation; 2) unimodal embedding; 3) multiscale visual self-attention; 4) visual-guided multimodal fusion (VGMF); and 5) optimized triplet loss.

### A. Formulation

First, given the features $v$ of the image $I$ and features $t$ of the text $T$, we need to map these two features to the space of dimension $d$ through two intramodal projections $E_v$ and $E_t$, which can be represented as $\hat{v} = E_v v$ and $\hat{t} = E_t t$ with $\hat{v} \in \mathbb{R}^{d_v}$, $\hat{t} \in \mathbb{R}^{d_t}$, where $d_v$ and $d_t$ represent the dimensions of visual features and text features, respectively. Then, the elementwise product is used to calculate the cross-modal similarity $\mathfrak{S}$ of the two embedding vectors $\hat{v}$ and $\hat{t}$

$$\mathfrak{S} = (W_{\hat{v}} \hat{v}) \odot (W_{\hat{t}} \hat{t}) \tag{1}$$

where $\odot$ denotes the elementwise product in matrices, $W_{\hat{v}} \in \mathbb{R}^{d_v \times d_f}$ and $W_{\hat{t}} \in \mathbb{R}^{d_t \times d_f}$ represent the features mapping matrix of embedded features $\hat{v}$ and $\hat{t}$, respectively, and $d_f$ represents the features dimension of the embedding space.

Furthermore, we have acquired a good mapping matrix $W_{\hat{v}}$, which can be used to obtain effective information in the RS image vector $\hat{v}$. During inference, the features representations of visual modal and text modal are independent, which indicates that the text encoding is not affected by visual information. However, when the text encoder considers the information of image and dynamically changes according to salient objects' information, the text representation turns out to be more reasonable. Therefore, it' is feasible to use image features to guide the text representation

$$\mathfrak{S} = (W_{\hat{v}} \hat{v}) \odot \Psi((W_{\hat{t}} \hat{t}), (W_{\hat{v}} \hat{v})) \tag{2}$$

where $\Psi(\mu, \lambda)$ implies that features $\lambda$ are used to guide the representation of features $\mu$.

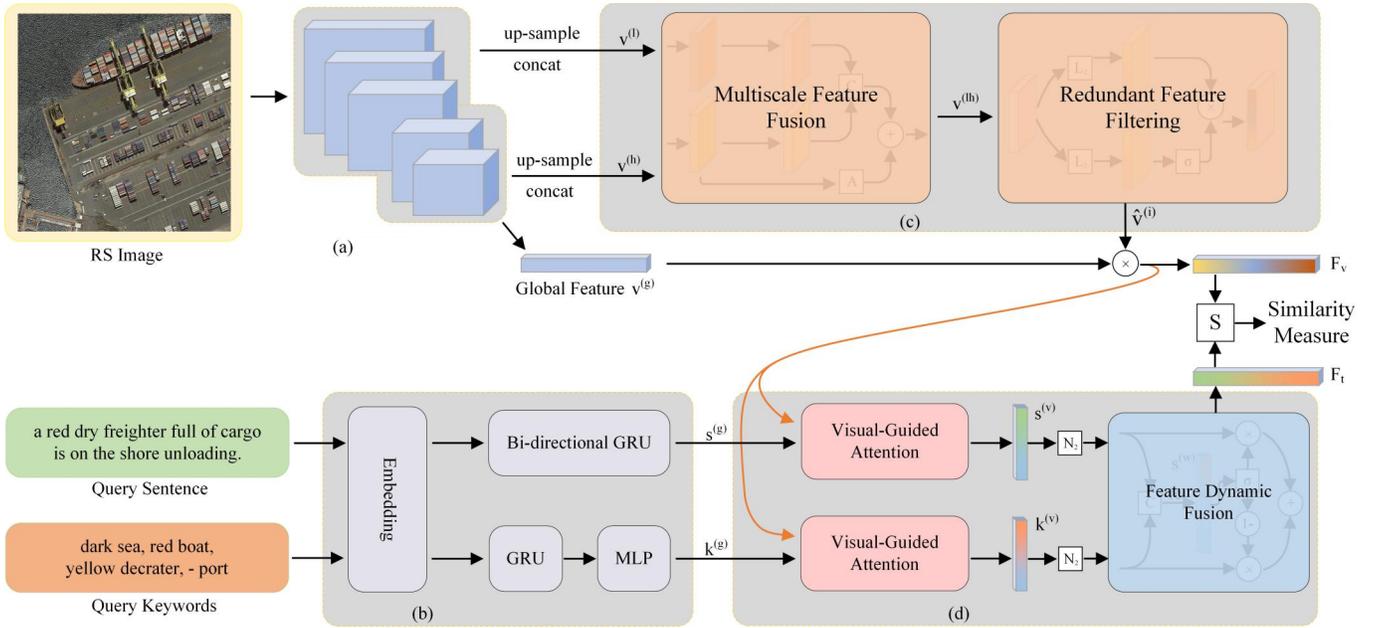

Fig. 2. AMFMN for RS image-text retrieval. AMFMN uses the MVSA module to obtain salient image features and uses salient features to guide the representation of text modalities. The network supports multiple retrieval methods and can adaptively fuse different modal text information. (a) CNN Module, (b) Text Embedding Module, (c) MVSA Module, (d) VGMF Module.

Keywords retrieval is usually applied to retrieve important objects in images, while sentence retrieval pays more attention to the adjacency relationship between multiple objects. In order to adapt the input to multiple retrieval methods (sentence and keywords), the model needs dynamically assign the weights for the input sentence and keywords to obtain robust text information representation, which can be written as

$$W_{\hat{t}}\hat{t} = W_{\hat{s}}\hat{s} \sqcup W_{\hat{k}}\hat{k} \qquad (3)$$

where $\sqcup$ denotes the dynamic fusion, and we define it as the dynamic weighted sum of two vectors. $\hat{s}$ and $\hat{k}$ represent the features of sentence and keywords, respectively, and $W_{\hat{s}}$ and $W_{\hat{k}}$ denote the representation matrix of sentence and keywords, respectively. Finally, the final retrieval formula can be expressed as

$$\mathfrak{S} = \left(W_{\hat{v}}\hat{v}\right) \odot \Psi\left(\left(W_{\hat{s}}\hat{s} \sqcup W_{\hat{k}}\hat{k}\right), \left(W_{\hat{v}}\hat{v}\right)\right). \qquad (4)$$

In Section III-B, we construct the image embedding matrix $E_v$, text embedding matrix $E_s$, and $E_k$ by using CNN and gate recurrent unit. Then, the MVSA module, which extracts the salient feature from RS image through image self-attention mechanism, is used to generate the image representation matrix $W_{\hat{v}}$. Furthermore, we construct the text representation matrices $W_{\hat{s}}$ and $W_{\hat{k}}$, design a dynamic fusion mechanism $\sqcup$, and utilize visual information to guide the dynamic integration of sentence features and keywords features. Finally, the positive sample ambiguity issue caused by the strong intraclass similarity in the RS image will be resolved by the proposed triplet loss function with a dynamic margin.

B. Unimodal Embedding

We will summarize the unimodal embedding part from three aspects: visual embedding, sentence embedding, and keywords embedding.

*1) Visual Embedding:* Consistent with the method in [36], for an input RS image, we utilize the CNN pretrained [38]–[40] on the ImageNet dataset and transfer its parameters to our task. Specifically, the global image features can be expressed as

$$v^{(g)} = \text{CNN}(I, \theta_I) \qquad (5)$$

where $I$ is an RS image, $v^{(g)}$ is the global features of the image extracted by CNN, and $\theta_I$ is used to represent the parameters of the CNN.

Although the feature $v^{(g)}$ contains most information of the RS image, there are apparent drawbacks for only using the global features $v^{(g)}$. On the one hand, the RS image contains a large number of targets compared with the natural image. Consequently, using global features $v^{(g)}$ directly for subsequent retrieval will cause information redundancy, which is unable to obtain favorable representations for the image. On the other hand, RS images have multiscale characteristics, and single-scale features cannot represent the image well. As the increase of convolution layer, some small targets in the image will be filtered out by the pooling layer, which means that these small targets will not appear in the global features. The deeper features map with a larger scale can capture high-level semantic information of salient objects, whereas the shallower features map with a smaller scale can extract fine features information. Considering the above factors, as shown in Fig. 2(a), we not only extract the global features from the RS image but also extract features from each layer of the convolution network.

After the extraction, motivated by the methods in [41], we upsample the feature maps of the first three layers and then concatenate these feature maps together as the low-level image features. Subsequently, the feature maps of the last two layers are sampled and connected as the high-level image features. The above process can be represented as

$$\{v^{(m)}\}_{m=1}^{5}, v^{(g)} = \text{CNN}(I, \theta_I) \quad (6)$$
$$\{F_m\}_{m=1}^{5} = \text{Upsample}(\{v^{(m)}\}_{m=1}^{5}) \quad (7)$$
$$v^{(l)} = \text{Cat}(F_1, F_2, F_3) \quad (8)$$
$$v^{(h)} = \text{Cat}(F_4, F_5) \quad (9)$$

where $v^{(m)}$ is each layer's feature map output by convolution network; the Upsample($\bullet$) operation denotes upsampled layer, which is used to match different feature maps to the same size. $F_m$ are the feature maps obtained after $v^{(m)}$ upsampled, where all feature maps of the low-level features have the same size, and the same for the high-level features. Cat($x, y$) is used to represent the channelwise concatenation of the feature vectors $x$ and $y$ with consistent dimensions. $v^{(l)}$ and $v^{(h)}$ represent the low-level features and high-level features extracted from RS image. Finally, we use the low-level features $v^{(l)}$, high-level features $v^{(h)}$, and global features $v^{(g)}$ as the final image features.

*2) Sentence Embedding:* For sentence, suppose that there is a sentence $S$, and we split it into $T$ words $\{w_t\}_{t=1}^{T}$. Then, each word is converted into a dense vector by the embedding matrix $W_e$, which is expressed by a formula $e_t = W_e(w_t)(t \in [1, T])$. Subsequently, considering the temporal information in the sentence, we use a bidirectional GRU to model the sentence and sequentially input the embedding words into the bidirectional GRU at different steps

$$h_t^f = \text{GRU}^f(e_t, h_{t-1}^f) \quad (10)$$
$$h_t^b = \text{GRU}^b(e_t, h_{t-1}^b) \quad (11)$$

where $h_t^f$ and $h_t^b$ represent the state of the forward GRU and the backward GRU at step $t$, respectively. We take the average of the hidden forward and backward states $s_t = (h_t^f + h_t^b/2)$ to represent the sentence feature vector at each step. As a result, the global information feature vector of the sentence can be represented by $s^{(g)}$

$$s^{(g)} = \frac{1}{T} \sum_{t=1}^{T} s_t. \quad (12)$$

*3) Keywords Embedding:* Similarly, keywords also need to be expressed with effective vectors. Assuming that a phrase $K$ containing $N$ keywords, the keywords can be recorded as $\{l_n\}_{n=1}^{N}$. Considering that keywords information is a supplement to sentence information, we make the keyword embedding matrix and sentence embedding matrix share the parameter matrix $W_e$, which is expressed by $j_n = W_e(l_n)(n \in [1, N])$. Even if all keywords are not relevant, since a single keywords phrase is usually two words (fine grained word and target word), a sequence model is still needed to model it. After that, we use multilayer perceptron (MLP) to get the final keyword features. Specifically speaking, we first input them into the GRU when characterizing individual keywords. The embedding process is consistent with sentence embedding. Finally, the global information feature vector of the keywords sequence can be expressed as $k^{(g)}$

$$k^{(g)} = \text{MLP}\left(\frac{1}{N} \sum_{n=1}^{N} (k_n)\right) \quad (13)$$

where $k_n$ is the GRU state when encoding the $n$th keyword.

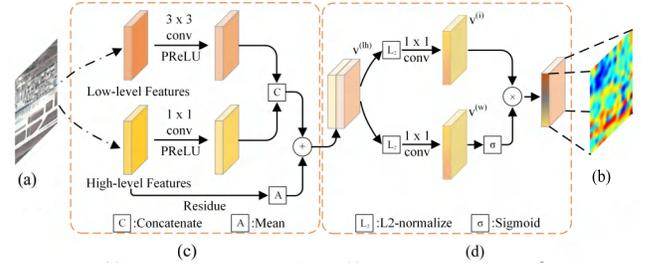

Fig. 3. MVSA. We first use a multiscale feature fusion network to obtain the multilevel feature representation, then use a redundant feature filtering network to filter out useless feature expressions, and finally get the salient mask of the RS image. (a) RS image. (b) Multiscale feature fusion. (c) Redundant feature filtering.

### C. Multiscale Visual Self-Attention

The motivation of the proposed MVSA module comes from the multiscale characteristics and the diversification of targets in RS image. An RS image usually contains abundant target information compared with natural images. For larger targets, the global feature vector can be used to express it better. However, for smaller targets, their target information in the RS image gradually decreases or even vanishes as the number of convolution networks increases. Besides, due to the complexity of RS image objects, the network needs to pay attention to the salient object information and suppress useless feature expression in the global features when retrieving by text.

In order to represent small targets in RS image and remove redundancy in high-level feature adaptively, the MVSA module is designed, as shown in Fig. 3. In short, for an RS image, we first use a multiscale feature fusion network to extract the joint information of each layer and then use a redundant target filtering network to adaptively filter the insignificant targets and extract the salient features of the image.

*1) Multiscale Feature Fusion:* Specifically, in the multiscale feature fusion stage, we perform feature transformations on the low- and high-level features. For low-level features, we use $3\times 3$ convolution layers to downsample them to the same feature map size as high-level features. After passing through the parametric rectified linear unit (PReLU) activation function, these features are used as the low-level feature representation $\widehat{v}^{(l)}$, which can be written as

$$\widehat{v}^{(l)} = \text{PReLU}(\text{conv}_{3\times 3}(v^{(l)})) \quad (14)$$

where PReLU($\bullet$) is the activation function, conv($\bullet$) is the convolution operation, and $\widehat{v}^{(l)}$ is the transformed low-level image features. Similarly, for high-level features $\widehat{v}^{(h)}$

$$\widehat{v}^{(h)} = \text{PReLU}(\text{conv}_{1\times 1}(v^{(h)})). \quad (15)$$

The purpose of using $3 \times 3$ and $1 \times 1$ convolution kernels is to make the two feature maps have the same size after transformation. Next, we concatenate the transformed high-level features $\widehat{v}^{(h)}$ and low-level features $\widehat{v}^{(l)}$ to obtain the joint representation. High-level features contain a more abstract expression of the RS image, which is very significant for global representation. In order to prevent the loss of high-level features, we take the mean value of the high-level features $v^{(h)}$ and connect it with the joint features by residual. Specifically

$$v^{(lh)} = \text{Mean}(v^{(h)}) \oplus \text{Cat}(\widehat{v}^{(l)}, \widehat{v}^{(h)}) \quad (16)$$

where $\text{Mean}(x)$ represents the mean of vector $x$, $v^{(lh)}$ is the joint feature representation of the high-level image features and the low-level image features, and $\oplus$ represents the vector summation.

*2) Redundant Feature Filter:* Although the fusion information of high- and low-level image features is obtained, there are still redundant feature representations in $v^{(lh)}$ due to the complexity of RS image targets. This redundant representation may cause visual features irrelevant to other representations and even produce misleading information. Naturally, the secondary feature transformation on the joint features is carried out to restrain the useless information of the joint features.

In the redundant feature filtering stage, we first perform the $L_2$ regularization on the joint features $v^{(lh)}$, and then, two independent $1 \times 1$ convolution operations are applied to generate the joint information features $v^{(i)}$ and the joint information feature gate vector $v^{(w)}$. Specifically

$$v^{(i)} = \text{conv}_{1\times1}(L_2(v^{(lh)})) \quad (17)$$
$$v^{(w)} = = \sigma(\text{conv}_{1\times1}(L_2(v^{(lh)}))) \quad (18)$$

where $L_2(x)$ represents the $L_2$ regularization of vector $x$ and $\sigma(x)$ represents the sigmoid activation function. Next, we exploit elementwise multiplication to suppress useless feature representations from the joint information features $v^{(i)}$ and obtain the final image information vector $\widehat{v}^{(i)}$

$$\widehat{v}^{(i)} = v^{(i)} \odot v^{(w)}. \quad (19)$$

After getting the image information gate vector $\widehat{v}^{(i)}$ and stretching it, we pass this vector to the sigmoid gate function to generate a salient mask of the global vector $v^{(g)}$. Then, we multiply $v^{(g)}$ and the salient mask to acquire the final image feature vector $F_v$.

### D. Visual-Guided Multimodal Fusion

The VGMF module is shown in Fig. 2(d), which uses visual information to guide the dynamic fusion of sentence features and keywords features. Even though the sentence used for the retrieval can reflect the adjacency attributes between multiple objects, it usually cannot comprehensively contain sufficient target information. The keywords retrieval can reflect the vital target in the RS image. Therefore, our motivation is to use the VGMF module to fuse sentence information and keywords information to obtain more robust and diverse feature expressions.

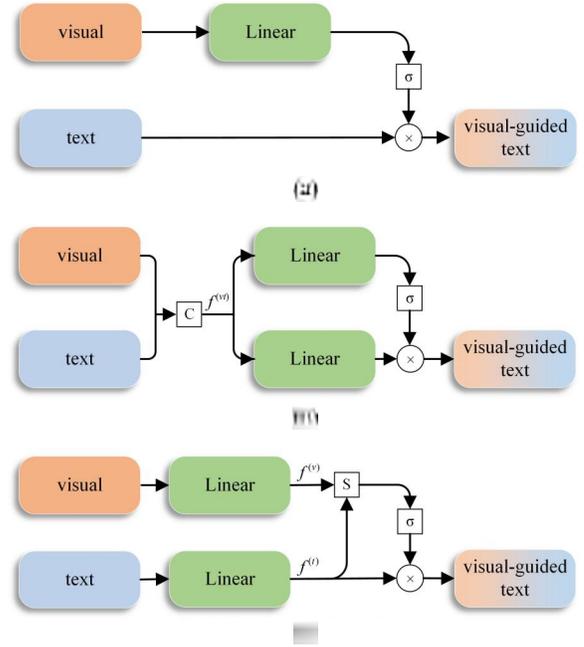

Fig. 4. Three different VGA mechanisms. (a) Soft attention. (b) Fusion attention. (c) Similarity attention.

*1) Visual-Guided Attention:* First, we design the visual-guided attention (VGA) module to obtain visual-guided information, which can provide guidance to other modal information. As shown in Fig. 4, three different VGA modules are proposed: soft attention, fusion attention, and similarity attention.

1) *Soft Attention:* Motivated by the method of Xu *et al.* [42], who uses visual information to generate gates through the linear layer to control the output of the text modal. Since the attention mechanism has become a classic approach, we take it as a baseline. After the visual information passes through the linear layer and the activation function, these features are directly used as the output gate of the text modal. Specifically

$$t^{(vg)} = \sigma(\text{linear}(v)) \odot t \quad (20)$$

where $v$ and $t$ are denoted as visual features and text features, respectively, $t^{(vg)}$ is the visual-guided text, and $\text{linear}(x)$ represents dealing features $x$ with linear transformation.

2) *Fusion Attention:* The motivation comes from attention on attention (AoA) [43]. A joint representation of visual modal and text modal is used to generate denoising self-attention. We connect visual features and text features in series to get the joint representation $f^{(vt)}$. After that, two identical linear layers are performed to generate text information and text information gates, respectively. In particular

$$f^{(vt)} = \text{Cat}(v, t) \quad (21)$$
$$t^{(vg)} = \text{linear}(f^{(vt)})) \odot \sigma(\text{linear}(f^{(vt)})). \quad (22)$$

3) *Similarity Attention:* We also learn the similarity calculation method of self-attention from [44] to generate

VGA. After the linear transformation of both visual information and text information, we measure the similarity between these two models by cosine similarity function. Then, a sigmoid function is applied to control the similarity and reform the focus on the converted text information. Specifically

$$f^{(v)} = \text{linear}(v) \tag{23}$$
$$f^{(t)} = \text{linear}(t) \tag{24}$$
$$t^{(vg)} = \sigma(S(f^{(v)}, f^{(t)})) \odot f^{(t)} \tag{25}$$

where $S(x, y)$ denotes the similarity between $x$ and $y$.

After the sentence and keywords features passing through the VGA module, the sentence features $s^{(v)}$ and the keywords features $k^{(v)}$ guided by visual information are obtained.

*2) Dynamic Fusion of Multimodal Information:* In addition to the visual-guided function, the other function of the VGMF module is the dynamic fusion of sentence information and keywords information. Some isolated target information in RS image is relatively rare in query sentence, whereas the information can be supplemented by keywords. We design a dynamic fusion method of multimodal information to enable sentence features and keywords features can fuse dynamically and effectively.

After adding $L_2$ regularization on the visual-guided sentence features $s^{(v)}$ and the visual-guided keywords features $k^{(v)}$, we concatenate these two sets of features in series. Subsequently, we use the neural network to generate the weighted vector $s^{(w)}$

$$s^{(w)} = \text{linear}(\text{Cat}(L_2(s^{(v)}), L_2(k^{(v)}))). \tag{26}$$

The activated $s^{(w)}$ is used as a dynamic weight function of $s^{(v)}$ and $s^{(k)}$, and the final representation $F_t$ of the text is formulated by

$$F_t = (s^{(v)} \odot \sigma(s^{(w)})) \oplus (s^{(k)} \odot (1 - \sigma(s^{(w)}))). \tag{27}$$

*E. Optimized Triplet Loss*

With the development of multimodal feature alignment, a triplet loss has been called one of the mainstream loss functions in the field of multimodal feature matching. The triplet loss aims to increase the distance between the sample and its corresponding negative samples and meanwhile make the distance between that sample and its positive sample as close as possible.

Faghri *et al.* [34] established a bidirectional triplet loss for text–image matching task

$$L(I, T) = \sum_{\widehat{T}}[\alpha - S(I, T) + S(I, \widehat{T})]_+$$
$$+ \sum_{\widehat{I}}[\alpha - S(I, T) + S(\widehat{I}, T)]_+ \tag{28}$$

where $\alpha$ represents the margin, $[x]_+ \equiv \max(x, 0)$, and $S(I, T)$ represent the similarity of image and text. The first sum is taken over all negative sentences $\widehat{T}$ given an image $I$, and the second sum considers all negative images $\widehat{I}$ given a sentence $T$. To improve the calculation efficiency, the loss

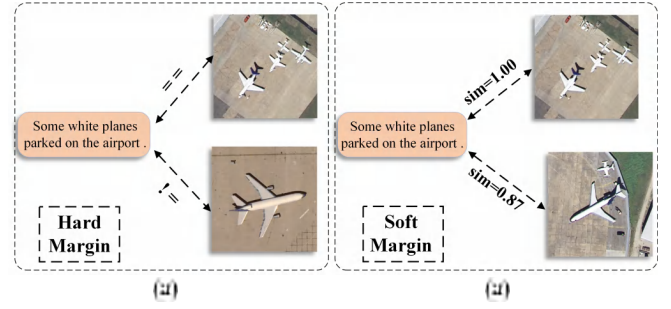

Fig. 5. Comparison of optimization goal between traditional methods and proposed methods. (a) Traditional optimization goal is to learn whether the image and text match, which is a two-classification problem. (b) Proposed optimization goal learns the degree of matching by using sample pairs' prior similarity.

is usually calculated in each batch rather than in all training sets.

However, in the traditional triplet loss calculation, for each negative sample, the margin is consistent. The situation is reasonable for natural images with more scene changes, but it is not suitable for RS images with lots of soft positive samples. In the left of Fig. 5, after learning the positive sample pair, the model will be ambiguous about the sample when it learns the corresponding soft samples. Due to the positive sample ambiguity problem, the model cannot learn a set of robust parameters from the data. On the other hand, the margin between two sample pairs in the same category should be smaller than the two different categories' margin. For this reason, we change the fixed margin in the triplet loss based on sample pairs' prior similarity to alleviate the positive sample ambiguity. As shown in Fig. 5, the optimization goal of the traditional model is to learn whether the image and text match, which is a two-classification problem, while we use sample pairs' prior similarity as the learning goal of the model, to learn the degree of matching between images and text. In short, we transform the hard label optimization problem into a soft label optimization problem.

Our purpose is to set proper margins for sample pairs with different similarities. For sample pairs that the image and text are completely unrelated, the margin between them should be as large as possible. On the contrary, for sample pairs with very similar images and text, the margin between them should be as small as possible. First, for sample pair $(T, I)$, we define $S(T, T_I) \in (0, 1)$ as the similarity between text $T$ and the five sentences corresponding to image $I$. During training, we use the BLEU and METEOR to calculate the text similarity in the training samples. The BLEU indicator is proposed in 2002 to evaluate the similarity of two texts by calculating the N-Gram count [45]. The METEOR indicator is based on the single-precision weighted harmonic average and the single word recall rate, and the result is highly correlated with the manual judgment [46].

Next, we define

$$\alpha_{\text{ct}} = \gamma \frac{-e^{\beta S(T, T_I)} + e^{\beta}}{-1 + e^{\beta}}, \quad \gamma \in (0, 1) \tag{29}$$

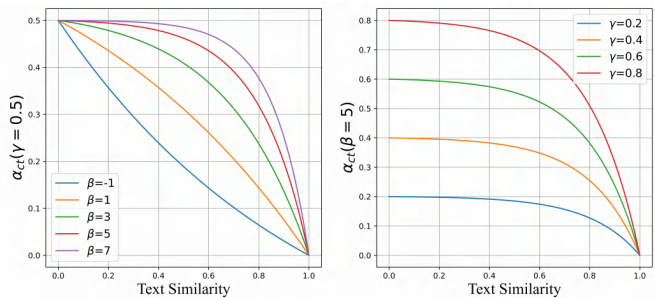

Fig. 6. Influence of parameters $\beta$ and $\gamma$ on $\alpha_{ct}$. (Left) Effect of $\beta$ change on $\alpha$ when $\gamma$ is 0.5. (Right) Effect of $\gamma$ change on $\alpha$ when $\beta$ is 5.

where $e$ is the natural index, $\gamma$ is the maximum margin, $\beta$ is the decay coefficient, and $\alpha_{ct}$ is the margin that has considered the text similarity. Fig. 6 shows the influence of parameters $\beta$ and $\gamma$ on $\alpha_{ct}$. As $\gamma$ gradually increases, the maximum margin of the curve gradually increases. When $\beta$ is greater than 0 and gradually increases, the curve is convex, the decline speed gradually becomes faster, and finally tends to zero when $S(T_I, T)$ tends to one. When $\beta$ is less than 0 and gradually decreases, the curve is concave. Finally, we define

$$L_{ct}(I, T) = \sum_{\widehat{T}}[\alpha_{ct} - S(I, T) + S(I, \widehat{T}_h)]_+ \\ + \sum_{\widehat{I}}[\alpha_{ct} - S(I, T) + S(\widehat{I}_h, T)]_+ \quad (30)$$

where $L_{ct}$ is the triplet loss with dynamic variable margin.

Formula (29) converts the prior similarity $S(T_I, T)$ into dynamic margin $\alpha_{ct}$. During training, the model will not only learn to distinguish positive and negative samples but also learn the degree of similarity between the negative and positive samples. Meanwhile, for RS images with strong intraclass similarity, the fixed margin $\alpha$ is converted into $\alpha_{ct}$, which greatly eliminates the positive sample ambiguity problem faced by the model.

## IV. DATASET FOR RS IMAGE-TEXT MATCH

This section first reviews the existing RS image-text dataset and then describes the proposed RSITMD.

### A. Existing Dataset in RS Image-Text Task

RS image caption datasets contain images and their corresponding sentences, which can be used directly if only functionality is considered. The primary image caption datasets of RS images include Sydney, UCM, and RSICD.

1) *Sydney-Captions Dataset:* Qu *et al.* [47] contributed to this dataset. The pixel resolution of the images is 0.5 m. The dataset contains 555 RS images, which are divided into seven scenes, including residential, airport, grassland, river, ocean, industrial area, and runway. Each image in the dataset is described by five sentences. Due to the limitation of capacity and high similarity, the dataset may be not enough to evaluate the superiority of text–image match model in RS.

2) *UCM-Captions Dataset:* This dataset is also contributed by Qu *et al.* [47]. The dataset contains a total of 2100 images of 21 scenes, each of which is $256 \times 256$ pixels. The images in the UCM dataset are manually extracted from urban area images of the National Map of the United States Geological Survey (USGS). The dataset also uses five captions to describe each image, but there is still the problem of high similarity between the sentences.

3) *RSICD-Captions Dataset:* The RSICD dataset is a large-scale RS image caption dataset. Considering the scale diversification and rotation invariance of RS images, Lu *et al.* [31] contributed the RS image dataset with a total of 10 921 images of 30 scenes. This dataset has become the preference for RS image caption tasks because of its large scale and various types. However, this dataset still cannot avoid the same demerits of repeated sentences (the text-to-image ratio, which we will be explained later, is 1.665), which causes the caption of plentiful similar scenes to be consistent and not more granular. Therefore, the RSICD dataset has the disadvantage of low fine graininess in RS image-text retrieval tasks.

To sum up, although there are already some datasets in the RS cross-modal retrieval field, these datasets may not be directly applicable to the RS image-text retrieval tasks due to the duplicated and imprecise text. To prove the above conclusions, we use the weighted sum of commonly used evaluation indicators in natural language processing (BLEU and METEOR) to evaluate the similarity of datasets and visualize the similarity of sentences in different datasets. For each text in the validation set, we use five texts corresponding to all images to calculate the similarity. The higher score represents higher similarity. The ideal curve is a straight diagonal line from the top left to the bottom right, which means that each sentence is only related to the corresponding image. After the calculation is completed, we draw the similarity results of each dataset, as shown in Fig. 7.

COCO and Flickr datasets are commonly used as image-text pair datasets in natural scenes. Fig. 7(a) and (b) shows the results of the similarity visualization of COCO and Flickr datasets, respectively. The similarity curves of the above two datasets are quite flat, which means that the language similarity in the dataset is extremely low. Fig. 7(d)–(f) shows the visualizations of the Sydney, UCM, and RSICD dataset. It can be seen that the intraclass similarity of these three datasets is very high, especially the Sydney dataset. The strong intraclass similarity means that one or more RS images may correspond to one sentence. Consequently, it may be unreasonable to train and evaluate the cross-modal RS image retrieval model using these three datasets.

This motivation drives us to propose a new dataset, RSITMD, suitable for RS image-text retrieval. Compared with the above datasets, our dataset has more fine-grained captions and minimizes the repetition between sentences. Fig. 7(c) shows the similarity visualization curve of RSITMD. Compared with the above three datasets, the similarity of RSITMD dataset is lower, and the similarity curve becomes a diagonal line. Meanwhile, we provide one to five fine-grained keywords for each sample image to further reduce the consistency of retrieval information, thereby promoting the development of RS image retrieval by sentence or keywords tasks.

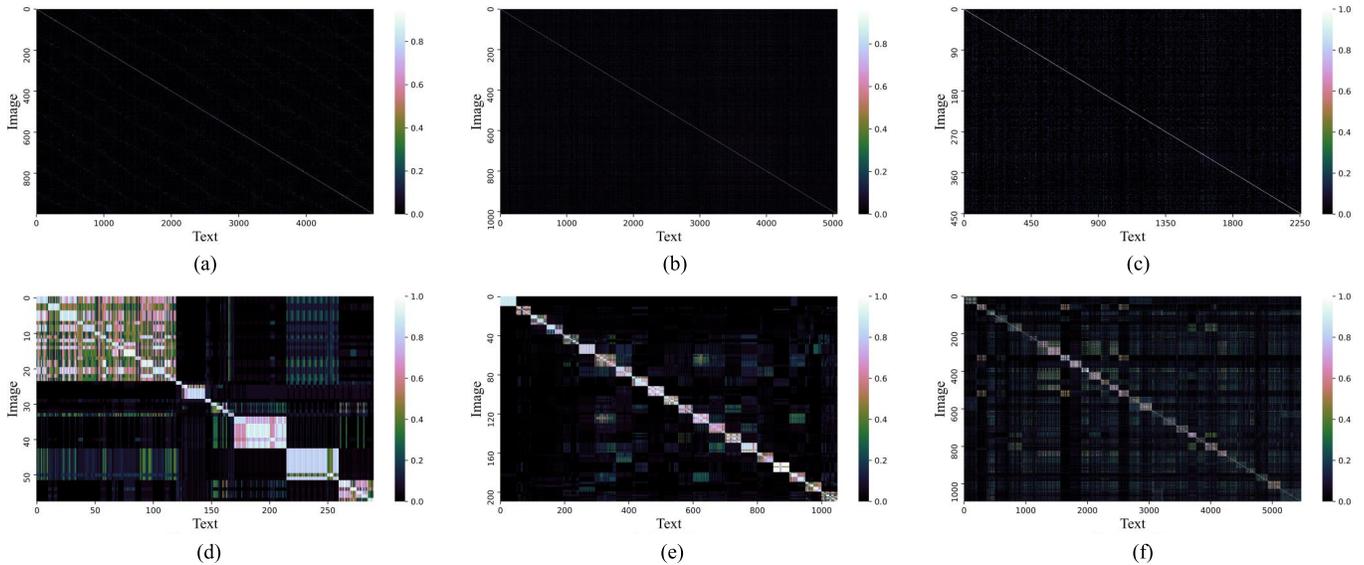

Fig. 7. Similarity visualization results of six datasets, where the similarity score is weighted by the BLEU and METEOR indicators in the natural language processing field. The ideal picture is a straight diagonal line from the top left to the bottom right, which means that each sentence is only related to the corresponding image. (a) COCO. (b) Flickr 30k. (c) RSITMD. (d) Sydney. (e) UCM. (f) RSICD.

TABLE I
DIFFERENT CLASS NUMBERS OF RSITMD

| Category | Number | Category | Number | Category | Number | Category | Number |
|---|---|---|---|---|---|---|---|
| industrial | 207 | port | 178 | beach | 154 | desert | 126 |
| stadium | 201 | farmland | 177 | commercial | 148 | forest | 116 |
| storagetanks | 201 | resort | 166 | center | 146 | railwaystation | 112 |
| square | 196 | school | 165 | parking | 143 | mountain | 109 |
| playground | 191 | park | 160 | airport | 140 | baseballfield | 105 |
| river | 188 | denseresidential | 157 | church | 136 | intersection | 67 |
| viaduct | 188 | sparseresidential | 157 | mediumresidential | 130 | bareland | 58 |
| pond | 184 | bridge | 155 | meadow | 127 | boat | 55 |

### B. RSITMD: A New Dataset for RS Image-Text Match

We construct a fine-grained RSITMD, which is used for RS cross-modal text–image retrieval. Some of the images in our dataset are selected from the RSICD dataset, while others are from Google Earth. We have provided a total of 23 715 captions for 4743 images, and the number of images in each category is shown in Table I.

When labeling, we require the annotators to provide information, such as the color, size, and adjacency of the target. Different annotators are also asked to provide captions for the same RS image. Since different annotators have inconsistent understanding of the same image, this approach can generate more diversified captions. Such a method can improve the characteristics of strong intraclass similarity and low interclass similarity of RS images and maximize the difference of intraclass images. We followed several principles when labeling:

1) The primary concern should be the fine-grained relationships between attributes of the objects in the image, not only the objects themselves.

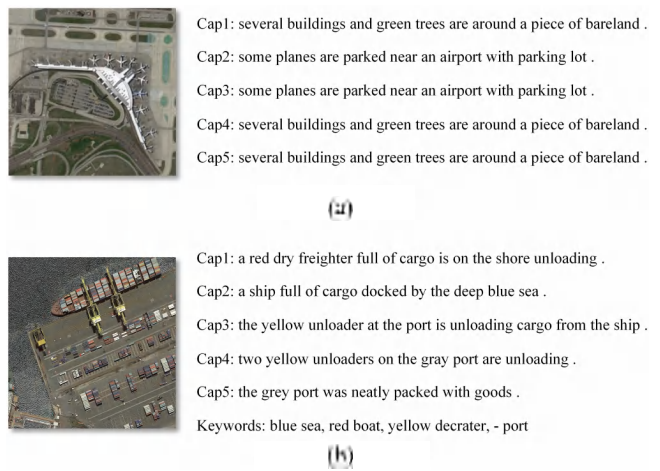

Fig. 8. Comparison of (a) RSICD and (b) RSITMD samples, each RSITMD sample contains five diversified descriptions and one to five fine-grained keywords.

2) Coarse words are used as little as possible in scenes that can be described by specific numbers, such as "some" and "many."
3) Try to use different words to express the same object, such as "ship" and "boat."

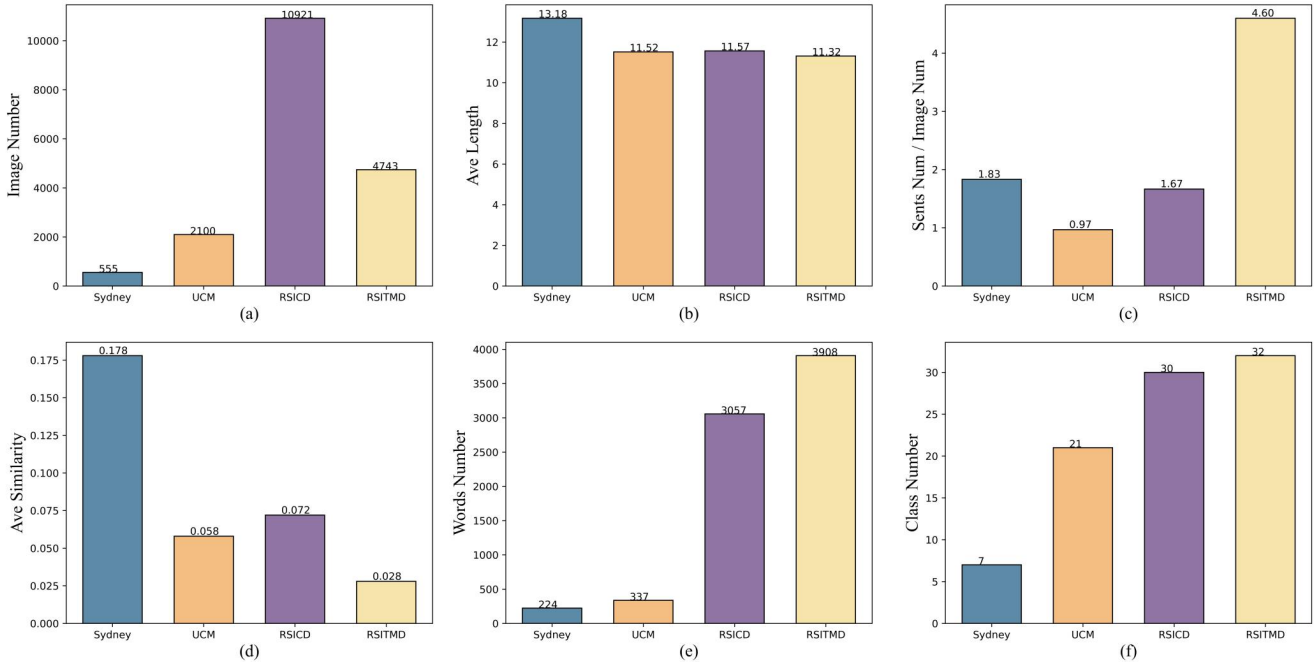

Fig. 9. Quantitative comparison of the four datasets. (a) Comparison of sample number. (b) Comparison of average sentence length. (c) Comparison of diversity score. (d) Comparison of average similarity. (e) Comparison of the total number of words. (f) Comparison of the number of categories.

4) When describing keywords, use "-" instead of fine-grained words for phrases without fine-grained information, such as "-port."

Fig. 8 shows the comparison between the sample of RSICD and RSITMD. RSITMD is more granular and it is more diverse in captions than RSICD.

The six-bar charts in Fig. 9 show a quantitative comparison of the four datasets. Although there are only 4743 images of RSITMD less than 10 921 images of RSICD, it is enough for the retrieval model to get robust parameters. We use the number of sentences that are completely inconsistent except for symbols in the dataset divided by the number of images to quantify the data diversity of different datasets. The quantization result is shown in the top right of Fig. 9. The diversity scores of Sydney, UCM, and RSICD datasets are 1.83, 0.97, and 1.67, respectively, and RSITMD's diversity score is 4.60, which proves that RSITMD has more different captions than others. Furthermore, we average the results of the visual similarity. The lower the score means, the lower the similarity of the dataset. We get the average similarity score, as shown in Fig. 9(d). Compared with other datasets, the similarity score of RSITMD is the lowest one. Meanwhile, even if RSITMD has fewer captions than RSICD, it contains more words and categories than RSICD as shown in Fig. 9(e) and (f), which once again proves the diversified characteristics of the proposed RSITMD dataset.

Retrieval of RS images by sentence and keywords is two different tasks. People often use a complete sentence to describe the adjacency relationship among multiple targets. For a single target, it is redundant to use a complete sentence to describe it. Therefore, even keywords retrieval methods and sentence retrieval methods both belong to text–image retrieval, they could have distinct requirements. We assign 1–5 keywords for each image in RSITMD so that this dataset can be applied to various RS retrieval methods. For example, RS images can be retrieved by using sentence or keywords individually or simultaneously. Keywords are used as additional information to supplement the target object in the RS image that does not appear in the sentence, thereby making the retrieval more accurate. We have equipped 4743 images with a total of 21 403 keywords to advance the task of joint retrieval of sentence and keywords in RS.

## V. EXPERIMENTS RESULTS AND ANALYSIS

We conducted a large number of experiments on the Sydney, UCM, RSICD, and RSITMD datasets to verify the effectiveness of the proposed AMFMN model on RS retrieval tasks.

### A. Dataset and Evaluation Metrics

For each dataset, we use 80% of the samples as the training set, 10% of the samples as the validation set, and the remaining 10% as the test set. We use the RSITMD dataset for the experiment of joint retrieval by keywords and sentence.

Two evaluation indicators $R@K (K = 1, 5, \text{and} 10)$ and mR are used to evaluate our model. $R@K$ represents the ratio of ground truth appearing in the top $K$ results. Besides, we also follow the average of all six recall rates of $R@K$ proposed by Huang *et al.* [48] to obtain mR, which is more reasonable for evaluating overall performance of the model.

### B. Implementation Details

All our experiments are carried out on a single NVIDIA Titan RTX GPU. For the image, we first scale it to $278 \times 278$

TABLE II
COMPARISONS OF SENTENCE-IMAGE RETRIEVAL EXPERIMENTS ON RSICD, RSITMD, UCM, AND SYDNEY TEST SET

| Approach | RSICD dataset | | | | | | | RSITMD dataset | | | | | | |
|---|---|---|---|---|---|---|---|---|---|---|---|---|---|---|
| | Sentence Retrieval | | | Image Retrieval | | | mR | Sentence Retrieval | | | Image Retrieval | | | mR |
| | R@1 | R@5 | R@10 | R@1 | R@5 | R@10 | | R@1 | R@5 | R@10 | R@1 | R@5 | R@10 | |
| VSE++ | 3.38 | 9.51 | 17.46 | 2.82 | 11.32 | 18.10 | 10.43 | 10.38 | 27.65 | 39.60 | 7.79 | 24.87 | 38.67 | 24.83 |
| SCAN t2i | 4.39 | 10.90 | 17.64 | 3.91 | 16.20 | 26.49 | 13.25 | 10.18 | 28.53 | 38.49 | 10.10 | 28.98 | 43.53 | 26.64 |
| SCAN i2t | **5.85** | 12.89 | 19.84 | 3.71 | 16.40 | 26.73 | 14.23 | 11.06 | 25.88 | 39.38 | 9.82 | 29.38 | 42.12 | 26.28 |
| CAMP-triplet | 5.12 | 12.89 | 21.12 | 4.15 | 15.23 | 27.81 | 14.39 | **11.73** | 26.99 | 38.05 | 8.27 | 27.79 | 44.34 | 26.20 |
| CAMP-bce | 4.20 | 10.24 | 15.45 | 2.72 | 12.76 | 22.89 | 11.38 | 9.07 | 23.01 | 33.19 | 5.22 | 23.32 | 38.36 | 22.03 |
| MTFN | 5.02 | 12.52 | 19.74 | 4.90 | 17.17 | 29.49 | 14.81 | 10.40 | 27.65 | 36.28 | 9.96 | 31.37 | 45.84 | 26.92 |
| AMFMN-soft | 5.05 | 14.53 | 21.57 | **5.05** | **19.74** | 31.04 | 16.02 | 11.06 | 25.88 | 39.82 | 9.82 | 33.94 | 51.90 | 28.74 |
| AMFMN-fusion | 5.39 | **15.08** | **23.40** | 4.90 | 18.28 | **31.44** | **16.42** | 11.06 | **29.20** | 38.72 | 9.96 | 34.03 | 52.96 | 29.32 |
| AMFMN-sim | 5.21 | 14.72 | 21.57 | 4.08 | 17.00 | 30.60 | 15.53 | 10.63 | 24.78 | **41.81** | **11.51** | **34.69** | **54.87** | **29.72** |
| Approach | UCM dataset | | | | | | | Sydney dataset | | | | | | |
| | Sentence Retrieval | | | Image Retrieval | | | mR | Sentence Retrieval | | | Image Retrieval | | | mR |
| | R@1 | R@5 | R@10 | R@1 | R@5 | R@10 | | R@1 | R@5 | R@10 | R@1 | R@5 | R@10 | |
| VSE++ | 12.38 | 44.76 | 65.71 | 10.10 | 31.80 | 56.85 | 36.93 | 24.14 | 53.45 | 67.24 | 6.21 | 33.56 | 51.03 | 39.27 |
| SCAN t2i | 14.29 | 45.71 | 67.62 | 12.76 | 50.38 | 77.24 | 44.67 | 18.97 | 51.72 | 74.14 | **17.59** | 56.90 | 76.21 | 49.26 |
| SCAN i2t | 12.85 | 47.14 | **69.52** | 12.48 | 46.86 | 71.71 | 43.43 | 20.69 | 55.17 | 67.24 | 15.52 | 57.59 | 76.21 | 48.74 |
| CAMP-triplet | 10.95 | 44.29 | 65.71 | 9.90 | 46.19 | 76.29 | 42.22 | 20.69 | 53.45 | **75.86** | 14.14 | 42.07 | 69.66 | 45.98 |
| CAMP-bce | 14.76 | 46.19 | 67.62 | 11.71 | 47.24 | 76.00 | 43.92 | 15.52 | 51.72 | 72.41 | 11.38 | 51.72 | 76.21 | 46.49 |
| MTFN | 10.47 | 47.62 | 64.29 | **14.19** | 52.38 | 78.95 | 44.65 | 20.69 | 51.72 | 68.97 | 13.79 | 55.51 | 77.59 | 48.05 |
| AMFMN-soft | 12.86 | **51.90** | 66.67 | **14.19** | 51.71 | 78.48 | 45.97 | 20.69 | 51.72 | 74.14 | 15.17 | 58.62 | 80.00 | 50.06 |
| AMFMN-fusion | **16.67** | 45.71 | 68.57 | 12.86 | **53.24** | **79.43** | **46.08** | 24.14 | 51.72 | **75.86** | 14.83 | 56.55 | 77.89 | 50.17 |
| AMFMN-sim | 14.76 | 49.52 | 68.10 | 13.43 | 51.81 | 76.48 | 45.68 | **29.31** | **58.62** | 67.24 | 13.45 | **60.00** | **81.72** | **51.72** |

pixels and then randomly crop and rotate the image to enhance the training samples. The final size of input images is 256×256 pixels. To guarantee the consistency of the experiment, while ensuring that the model will not be affected by the overfitting caused by the deep network, we use ResNet-18 [49] as the visual feature extractor, and the visual embedding dimension size is 512. The word embedding dimension is set to 300, and the hidden layer of the bidirectional GRU is set to 512. We use the Adam optimizer to train the network for 150 epochs, and the minimum batch size is set to 128. During training, the learning rate is adjusted to $1e^{-4}$, and the learning rate is declined by 0.7 every 20 epochs.

### C. Comparisons With the State-of-the-Art Approaches

In this experiment, we contrast the AMFMN method with the following four methods on four RS image-text datasets.

1) *VSE++ [34]:* VSE++ is one of the pioneers of image retrieval in the field of natural images. The author uses convolution networks and recurrent networks to embed image information and text information into the same space and proposes the triplet loss to train the image-text matching model.
2) *SCAN [35]:* The SCAN model, which is based on VSE++, uses Faster RCNN [50] to extract image features and tries to align the target in the image with the target in the text. In this experiment, we compared with the t2i method and i2t method proposed in this article.
3) *CAMP [51]:* The CAMP model proposes a method of adaptive message passing, which adaptively controls the flow of cross-modal information transmission and calculates the matching scores of images and texts by using fusion features. The triplet loss method and the bce loss method of CAMP are used as control groups.
4) *MTFN [36]:* The MTFN model designs a multimodal fusion network based on the idea of rank decomposition to calculate the distance of the embedding features.

Among the above methods, we use the ResNet-18 network to replace the object detection network. Since the proposed AMFMN needs to be inputted with both keywords and sentence, we filter the keywords in sentence and input them into keywords branch when only the sentence is input. When only the keywords are input, we connect the keywords and use them as the input of the sentence branch. Table II shows the test results of the model on four datasets: RSICD, RSITMD, UCM, and Sydney.

*1) Results on RSICD:* The comparison of models on the RSICD dataset is in the top-left area of Table II. We show the results of the model when using three different VGA modules: AMFMN-soft, AMFMN-fusion, and AMFMN-sim. No matter which VGA module we use, our experiment scores are better than other models. In other words, our AMFMN-fusion model is the best among all models. Compared with the MTFN that uses the fusion method for matching, our experiment results have improved mR by 1.95, which proves the superiority of our method.

*2) Results on RSITMD:* The results on the RSITMD dataset are in the top-right area of Table II. On the RSITMD dataset, except for the single indicator *R*@1 of sentence recall, our model achieves the best results for other indicators. Specially speaking, AMFMN-sim has achieved all first in the image recall indicators of the RSITMD dataset, and it achieves the first place in comprehensive performance. The combined

TABLE III
COMPARISONS OF KEYWORDS-IMAGE RETRIEVAL EXPERIMENTS ON THE RSITMD TEST SET

| Approach | Keywords Retrieval | | | Image Retrieval | | | mR |
|---|---|---|---|---|---|---|---|
| | R@1 | R@5 | R@10 | R@1 | R@5 | R@10 | |
| VSE++ | 2.43 | 8.63 | 13.27 | 5.31 | 18.54 | 29.38 | 12.93 |
| SCAN | **7.96** | 10.84 | 18.14 | 8.19 | 24.25 | 36.24 | 17.60 |
| CAMP | 6.19 | 12.39 | 18.58 | 5.85 | 21.90 | 33.41 | 16.39 |
| MTFN | 6.19 | 6.19 | 11.73 | 7.08 | **29.20** | **46.24** | 17.77 |
| AMFMN | 7.52 | **18.14** | **24.34** | **8.27** | 26.41 | 40.13 | **20.80** |

TABLE IV
COMPARISONS OF USING MULTISOURCE INPUT TO RETRIEVAL RS IMAGES ON THE RSITMD TEST SET

| Input | Text Retrieval | | | Image Retrieval | | | mR |
|---|---|---|---|---|---|---|---|
| | R@1 | R@5 | R@10 | R@1 | R@5 | R@10 | |
| Sentence | 10.63 | 24.78 | **41.81** | 11.51 | 34.69 | **54.87** | 29.72 |
| Keywords | 7.52 | 18.14 | 24.34 | 8.27 | 26.41 | 40.13 | 20.80 |
| Sentence & Keywords | **11.06** | **28.76** | 40.27 | **12.30** | **36.68** | 54.38 | **30.58** |

scores of the other two models are also far superior to other models.

*3) Results on UCM:* The bottom left of Table II shows the scores of each model on the UCM dataset. Although the SCAN i2t method achieves the best $R@10$ indicator in the sentence recall indicator, it is still lower than AMFMN in most indicators. The comprehensive performance of the AMFMN-soft method is also very prominent, in which the $R@5$ indicator of sentence recall and the $R@1$ indicator of image recall have reached the best.

*4) Results on Sydney:* The Sydney dataset challenges the robustness of the model due to its small size. The bottom-right area of Table II shows the results on the Sydney dataset. The AMFMN-soft model and AMFMN-fusion model have achieved the same effect and are tied for second place. The overall performance of the proposed AMFMN-sim ranks first, with the highest score of 51.72 on the mR indicator. The performance on the Sydney dataset strongly illustrates the robustness of the AMFMN method.

### D. Retrieval Images With Keywords

The RSITMD dataset contains 21 403 keywords, and we use this dataset for the task of keywords-image retrieval. For other models, we connect the keywords and input them into the model for comparison. The results of keywords-image retrieval are shown in Table III. For this retrieval task, the accuracy of AMFMN is optimal, and the score on mR indicator is 20.80, which is higher than MTFN with a score of 3.03 points.

In RS image retrieval, a complete sentence usually describes multiple objects and the relative adjacency relationship. However, such a complete sentence usually cannot summarize all the information of the RS image. Therefore, if the keywords information in the RS image is used to supplement the sentence information, higher retrieval accuracy will be obtained.

For example, we add missed target information of the RS image descriptions, such as "dark sea" not mentioned in the query sentence in Fig. 2 and let the input keywords information supplement the information of sentence. We experiment with this idea to verify whether this method can improve the retrieval accuracy.

First, we only use the sentence to retrieve RS images, then use keywords to retrieve RS images, and finally use the combination of both to retrieve RS images. We use all the keywords information when performing keywords-image retrieval on the RSITMD dataset. In the joint retrieval of sentence and keywords, keywords tend to have a strong intraclass similarity effect on the retrieval results. In order to reduce the impact, we remove the words that frequently appeared in the keywords during the training process and obtain more fine-grained keywords.

The experiment results are shown in Table IV. Compared with sentence-image retrieval, the method of keywords-image retrieval is less effective. The reason is that the strong intraclass similarity of keywords makes the features of objects more consistent. Consequently, it is still necessary to use fine-grained sentence features to distinguish and retrieve. For the retrieval method of input sentence and keywords simultaneously, the effect is improved by 0.86 points compared with the method of sentence-image retrieval. It is proved that keywords retrieval can supplement more information into the query sentence to further improve retrieval accuracy.

### E. Ablation Studies

In this section, we will carry out ablation experiments on the proposed model and systematically explore the influence of each module on the experiment results.

We experiment with the dynamic variable triplet loss function ($L_{\text{ct}}$) in advance. We set up three groups of constant margins to observe the influence of $L_{\text{ct}}$ on the experiment results. At the same time, in order to observe the influence of $\gamma$ and $\beta$ parameters on the $L_{\text{ct}}$, we also control these two parameters to observe the recall rate. The experiment results are shown in Table V.

The notation $\gamma$ represents the maximum margin. If there is no data with strong similarity, $L_{\text{ct}}$ of $\gamma = 0.4$ will degenerate into traditional triplet loss function ($L$) of $\alpha = 0.4$. First, we compare the effect of changing $\beta$ on the experiment results under the same maximum margin. When the maximum margin is 0.4, the mR indicator of $L_{ct}$ is higher than that of $L$ regardless of the value of $\beta$. The red font in Table V represents the maximum mR for different loss functions $L$ and $L_{ct}$ when the margin is constant. Similarly, this conclusion can also be drawn when the maximum margin is 0.6 or 0.8. When we fix $\gamma = 0.4$ and change $\beta$, we find that the recall rate is lowest when $\beta = 3$. Similarly, when $\gamma = 0.6$ or $\gamma = 0.8$, this conclusion is also found, but the magnitude of the decrease in mR is not significant. When observing the influence of other $\beta$ on the experiment results, it can be found that parameter $\beta$ is insensitive to the indicator mR. Therefore, we can arrive at the following conclusions; when the maximum margin changes, $L_{ct}$ performs better than $L$, and the parameter $\beta$ has no significant effect on the final experiment results.

TABLE V
EXPERIMENT OF LOSS FUNCTION CONTROL ON SENTENCE-IMAGE RETRIEVAL

| Loss Function | | | Sentence Retrieval R@1 | R@5 | R@10 | Image Retrieval R@1 | R@5 | R@10 | mR |
|---|---|---|---|---|---|---|---|---|---|
| $triplet$ | $\alpha = 0.4$ | | 11.50 | 26.11 | 38.05 | 10.49 | 32.65 | 48.41 | 27.87 |
| | $\alpha = 0.6$ | | **14.82** | **28.98** | 37.61 | 9.12 | 31.68 | 46.90 | 28.19 |
| | $\alpha = 0.8$ | | 12.39 | 27.88 | 40.49 | 10.49 | 32.65 | 46.86 | 28.46 |
| $triplet_{ct}$ | $\gamma = 0.4$ | $\beta = 3$ | 9.29 | 26.11 | 40.93 | 9.47 | 34.47 | 52.65 | 28.82 |
| | | $\beta = 5$ | 12.17 | 26.77 | 38.94 | 11.02 | 35.97 | 52.43 | <span style="color:red">29.55</span> |
| | | $\beta = 7$ | 11.50 | 26.55 | 40.04 | 10.97 | 34.25 | 51.46 | 29.13 |
| | $\gamma = 0.6$ | $\beta = 3$ | 10.18 | 25.88 | 38.94 | 9.56 | 34.42 | 52.26 | 28.54 |
| | | $\beta = 5$ | 10.63 | 24.78 | **41.81** | 11.51 | 34.69 | **54.87** | **29.72** |
| | | $\beta = 7$ | 11.28 | 27.21 | 38.05 | 11.28 | 35.27 | 52.08 | 29.20 |
| | $\gamma = 0.8$ | $\beta = 3$ | 8.63 | 25.22 | 39.16 | 10.71 | 35.71 | 53.41 | 28.81 |
| | | $\beta = 5$ | 8.41 | 24.78 | 40.04 | **11.64** | **36.99** | 54.60 | 29.41 |
| | | $\beta = 7$ | 8.41 | 28.10 | 40.49 | 10.88 | 36.37 | 52.92 | <span style="color:red">29.53</span> |

TABLE VI
EXPERIMENTAL RESULTS OF AFMMN MODELS WITH DIFFERENT CONFIGURATIONS

| | Ablation Model | Visual $v^{(g)}$ | $F_v$ | Text $s^{(g)}$ | $k^{(g)}$ | $s^{(v)}$ | $k^{(v)}$ | $F_t$ | Text Retrieval R@1 | R@5 | R@10 | Image Retrieval R@1 | R@5 | R@10 | mR |
|---|---|---|---|---|---|---|---|---|---|---|---|---|---|---|---|
| baseline | m1 | ✓ | | ✓ | | | | | 10.38 | 27.65 | 39.60 | 7.79 | 24.87 | 38.67 | 24.83 |
| | m2 | ✓ | | | ✓ | | | | 5.75 | 17.26 | 25.56 | 4.69 | 21.33 | 33.23 | 17.97 |
| MVSA | m3 | | ✓ | ✓ | | | | | 10.95 | 25.22 | 35.84 | 9.78 | 31.06 | 46.19 | 26.51 |
| | m4 | | ✓ | | ✓ | | | | 7.30 | 17.26 | 24.56 | 5.75 | 23.54 | 35.40 | 18.97 |
| | m5 | | ✓ | ✓ | ✓ | | | | 10.94 | **28.98** | **44.91** | 9.69 | 32.92 | 46.28 | 28.95 |
| VGMF | m6 | | ✓ | | | ✓ | | | 10.63 | 24.78 | 41.81 | 11.51 | 34.69 | **54.87** | 29.72 |
| | m7 | | ✓ | | | | ✓ | | 7.52 | 18.14 | 24.34 | 8.27 | 26.41 | 40.13 | 20.80 |
| | m8 | | ✓ | | | ✓ | ✓ | | 10.17 | 28.10 | 40.93 | 9.87 | 35.31 | 54.03 | 29.74 |
| AMFMN | m9 | | ✓ | | | | | ✓ | **11.06** | 28.76 | 40.27 | **12.30** | **36.68** | 54.38 | **30.58** |

To better analyze the performance of the model, we list a series of configurations in Table VI. For visual representation, we use $v^{(g)}$ and $F_v$ to represent the global image features and the features extracted by the MVSA module, respectively. For text representation, we use $s^{(g)}$ and $k^{(g)}$ to represent the global features of the sentence and the keywords, $s^{(v)}$ and $k^{(v)}$ to represent sentence features and keywords features of visual guidance, and $F_t$ to represent visual-guided text features.

In order to explore the performance of each module effectively, we carefully set up nine ablation experiments:

1) *m1($v^{(g)} + s^{(g)}$):* Only measure the similarity between image global features and sentence global features.
2) *m2($v^{(g)} + k^{(g)}$):* Only measure the similarity between image global features and keywords global features.
3) *m3($F_v + s^{(g)}$):* Measure the similarity of visual features extracted by the MVSA module and sentence global features.
4) *m4($F_v + k^{(g)}$):* Measure the similarity of visual features extracted by the MVSA module and keywords global features.
5) *m5($F_v + s^{(g)} + k^{(g)}$):* Measure the similarity of visual features extracted by the MVSA module and the weighted sum of sentence global features and keywords global features.
6) *m6($F_v + s^{(v)}$):* Measure the similarity of visual features extracted by the MVSA module and sentence features extracted by the VGA module.
7) *m7($F_v + k^{(v)}$):* Measure the similarity of visual features extracted by the MVSA module and keywords features extracted by the VGA module.
8) *m8($F_v + s^{(v)} + k^{(v)}$):* Measure the similarity of visual features extracted by MVSA module, and the weighted sum of sentence features and keywords features extracted by the VGA module.

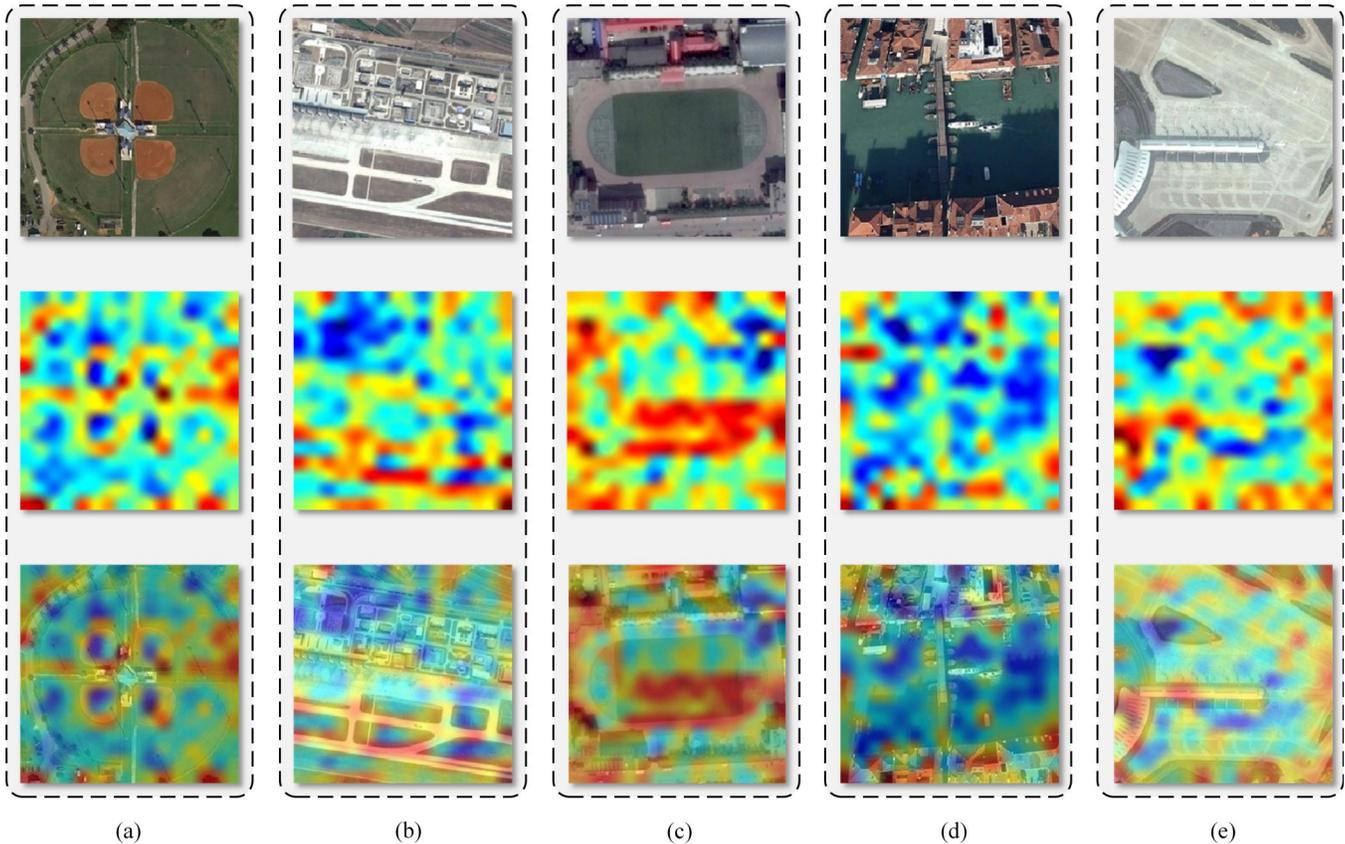

Fig. 10. Visualization of salient mask output by the MVSA module. (a)–(e) respectively represent the visualization results of five different samples. (Top) Original image. (Middle) Salient mask. (Bottom) Superimposed image.

9) $m9(F_v + F_t)$: Measure the similarity between visual features extracted by the MVSA module and dynamic fusion features extracted by the VGMF module.

Among them, m1 and m2 aims to provide a benchmark for the experiment; m3–m5 are control experiment of the MVSA module, aiming to observe the change in retrieval accuracy after changing the $v_{(g)}$ to $F_v$; m6–m8 are the comparison of the VGMF module, focusing on exploring the validity of sentence features and keywords features at different stages; and m9 shows the results of RS image retrieval using all the components.

Table VI shows the text–image retrieval results of AMFMN and its variants on the RSITMD dataset.

1) Comparing m1 with m2, the keywords retrieval has a lower accuracy than sentence retrieval. Similarly, when the image features are all $F_v$, the accuracy of keywords retrieval is still lower than that of sentence retrieval, which proves that the model can obtain more fine-grained information by using sentence.
2) Comparing m1 with m3, when the image features are changed from $v^{(g)}$ to $F_v$, the retrieval indicator has been improved by 1.18 points. By comparing m2 with m4, we can reach the same conclusion.
3) When paying attention to m3–m5, it can be found that when we use different retrieval methods, the retrieval accuracy will be improved accordingly.
4) Comparing m3 with m6 or m4 with m7, for the same image features, when the text module uses the VGA mechanism, the results also have corresponding improvements.
5) By comparing m8 with m9, the performance of the model has been improved correspondingly after adding the adaptive gate, which proves the effectiveness of the proposed information fusion mechanism.

*F. Saliency Mask Visualization*

In this section, we visualize the salient region mask to analyze the function of the MVSA module. The salient mask enables the network to extract salient features of the image and adaptively analyze the query statements focusing on which areas of the image.

The salient mask of five typical images is shown in Fig. 10. In Fig. 10(a), people are most likely to describe it as "four tennis courts are next to each other," so the model pays more attention to the four red tennis courts in the corresponding salient mask. At the same time, the model has also noticed attention to the gray roads, indicating that the model has learned to focus on some more fine-grained information. In Fig. 10(b), the model emphasizes the runway part of the airport, representing that the algorithm has concluded from the dataset that people prefer to describe the relationship between the runway and lawn in the airport. In Fig. 10(c), we can see that the model focuses on the green playground and red runway and also pays more attention to the red houses beside the playground. This sample fully proves that the model thinks that when describing the image, it is easier for people to start with the adjacent relationship between the playground and the runway and between the playground and the adjacent red chamber. In Fig. 10(d), it is strange that the model focuses on the houses on both sides of the river and the ships in the river rather than the bridge. The reason could be that the algorithm considers the boats in the river and the boats on both sides of the river are more representative than this bridge. In Fig. 10(e), the model highlights on the airport buildings and lawns, which is consistent with the labeled caption. These experiments are sufficient to prove that the MVSA module can effectively focus on the salient objects of the RS image.

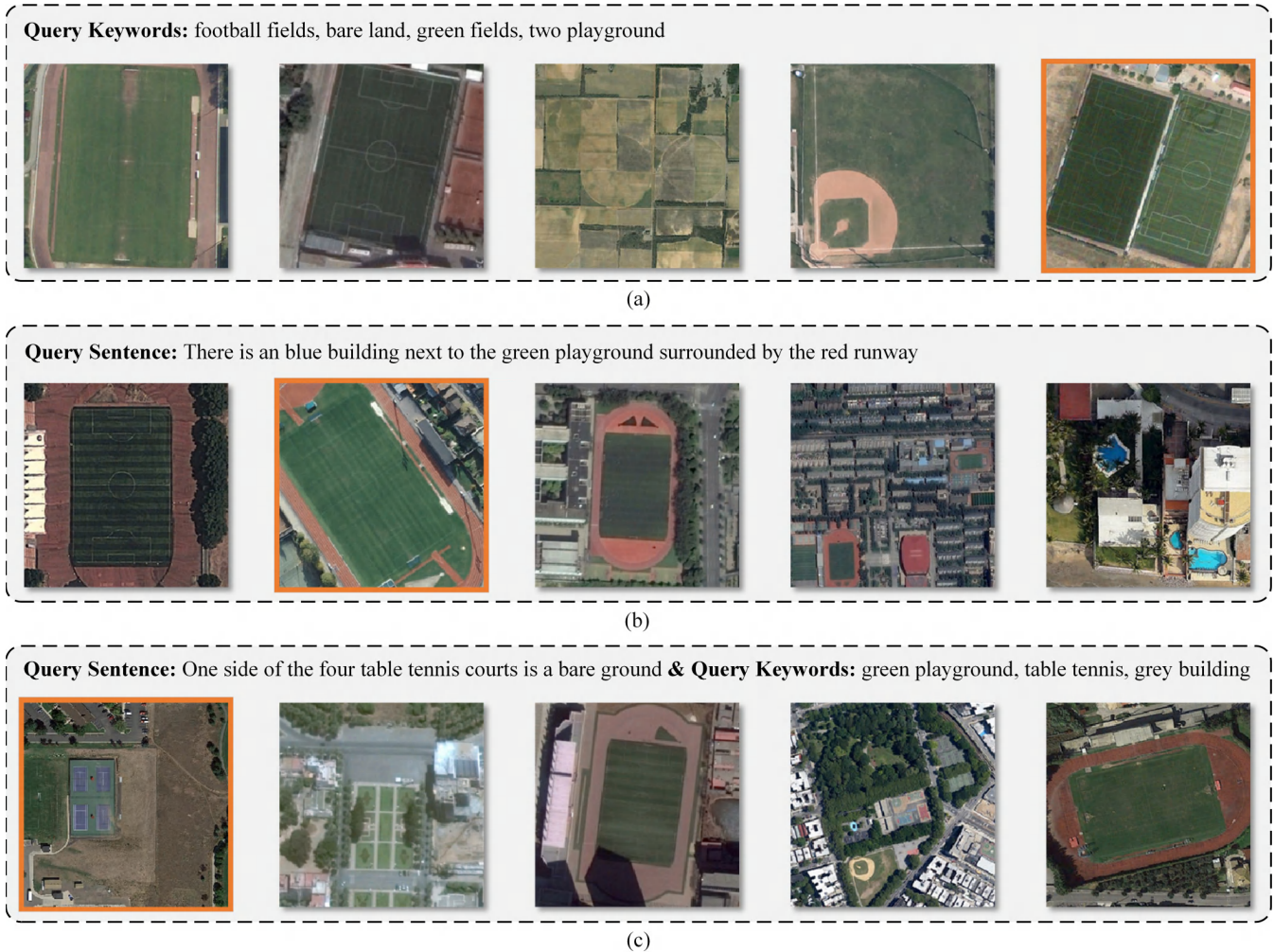

Fig. 11. Some of the results of retrieval samples. (a) Results of keywords retrieval. (b) Results of sentence retrieval. (c) Combination of sentence and keywords used for retrieval. (From Left to Right) They are ranked 1–5, and the image in the orange box is the ground truth.

### G. RS Image Recall Analysis

In this section, as in Fig. 11, we show some results of image retrieval by sentence and keywords and analyze the advantages and disadvantages of both. We merely select some of the typical results as samples for analysis. The result in the orange box is the ground truth corresponding to each query text.

When retrieving only by keywords, the result is shown at the top of Fig. 11(a). Even if the keywords do not contain fine-grained expressions, this method can still achieve good results. It should be noted that although first a few results are wrong, the wrong results still belong to the category of the ground truth. For example, when performing keywords retrieval, rank1 and rank2 retrieved still belong to the category of the playground. For the keywords retrieval, even though the recall rate is low, the category of the wrong samples is consistent with the ground truth.

Next, the result of sentence-image retrieval is shown in the middle of Fig. 11(b). Compared with keywords retrieval, sentence retrieval can already get considerable results. Rank 1 corresponding to sentence retrieval is not the ground truth, but they are very similar, which indicates that higher retrieval accuracy can be obtained by using sentence. However, for Rank5 by sentence retrieval, the model regards the red roof as a playground, and the corresponding category is far from the ground truth. Therefore, compared with the keywords retrieval, the results of sentence retrieval are more likely to be mismatched with the ground truth category. Finally, for joint retrieval using sentence and keywords, the advantages of both are obtained. On the one hand, the combination of keywords and sentences can achieve higher retrieval accuracy. On the other hand, the category of error results is consistent with that of correct RS image. The above experiments also prove that higher retrieval accuracy can be obtained by using more fine-grained information.

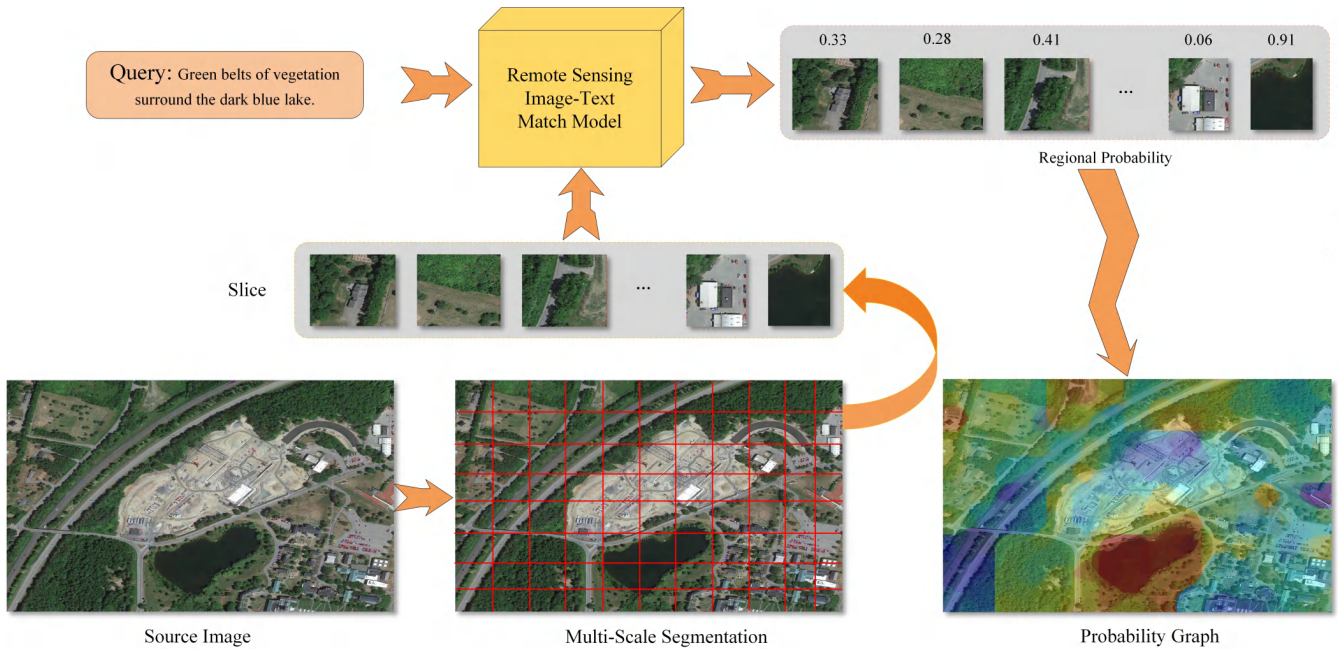

Fig. 12. Flowchart of location by using text.

*H. Explore Fast Locate Using Text*

In this section, we attempt to use the text to locate RS images in large scenes. The overall flowchart of the experiment is shown in Fig. 12. To maintain the integrity of the target, the large-scale RS image is segmented according to different scales and offsets. Then, we calculate the similarity between each RS slice and the query text and concatenate these similarities at the pixel level. Finally, we used a median filter to filter out the impulse noise of the probability map to obtain a more robust result.

First of all, splitting the RS image only once will inevitably result in the incompleteness of the target. Besides, when we locate RS images, due to the lack of corresponding scale information, the target may be too small or too large if only single-scale RS slices are used. Based on this fact, we attempt to slice RS images several times within the acceptable operating efficiency range, so as to obtain RS slices of different scales. Specifically, we set the slice size to $256 \times 256$ pixels. After the first cut, as shown in Fig. 13, we move the sliding window to the bottom right by $1/2$ grid and then cut the RS image again. The result generated by the above cutting method is considered as the result of the first round of cutting. Furthermore, we change the size of slice and repeat the above steps with grids of $128 \times 128$ pixels and $512 \times 512$ pixels,

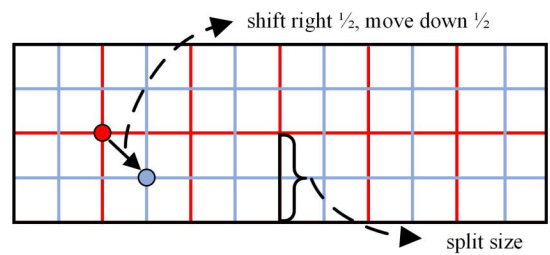

Fig. 13. Multiscale segmentation method. After the first cutting, we move the vertices of the cutting box by 1/2 grid and regard the above operation as a round. Then, we changed the size of the slice and cut it with three rounds.

respectively. Finally, different scales of RS image slices are obtained.

Next, AMFMN is used to predict the similarity between each RS slice and the query text, and the multilevel similarities are averaged to get the probability map in the large-scale RS scene. To filter out the impact signal in the probability map, we perform median filtering on the generated probability map, so as to obtain the final location result.

We show some of the representative location results, as shown in Fig. 14. In Fig. 14(a), we try to find six tennis courts adjacent to the trees from the large scene image. Even if the generated probability map can locate the ground truth, part of the probability still falls on some lawn and trees, which reflects that the model has room for improvement. In Fig. 14(b), we attempt to find a gray rectangular house beside the white roof and the model has been successfully focused on the right region. In Fig. 14(c), we plan to retrieve a highway with trees on one side and bare land on the other. Even though the performance of the model is not as accurate as results of road extraction, it can still roughly locate the road described. Compared with Fig. 14(d), we intend to retrieve for more fine-grained roads. Although the residential area has received

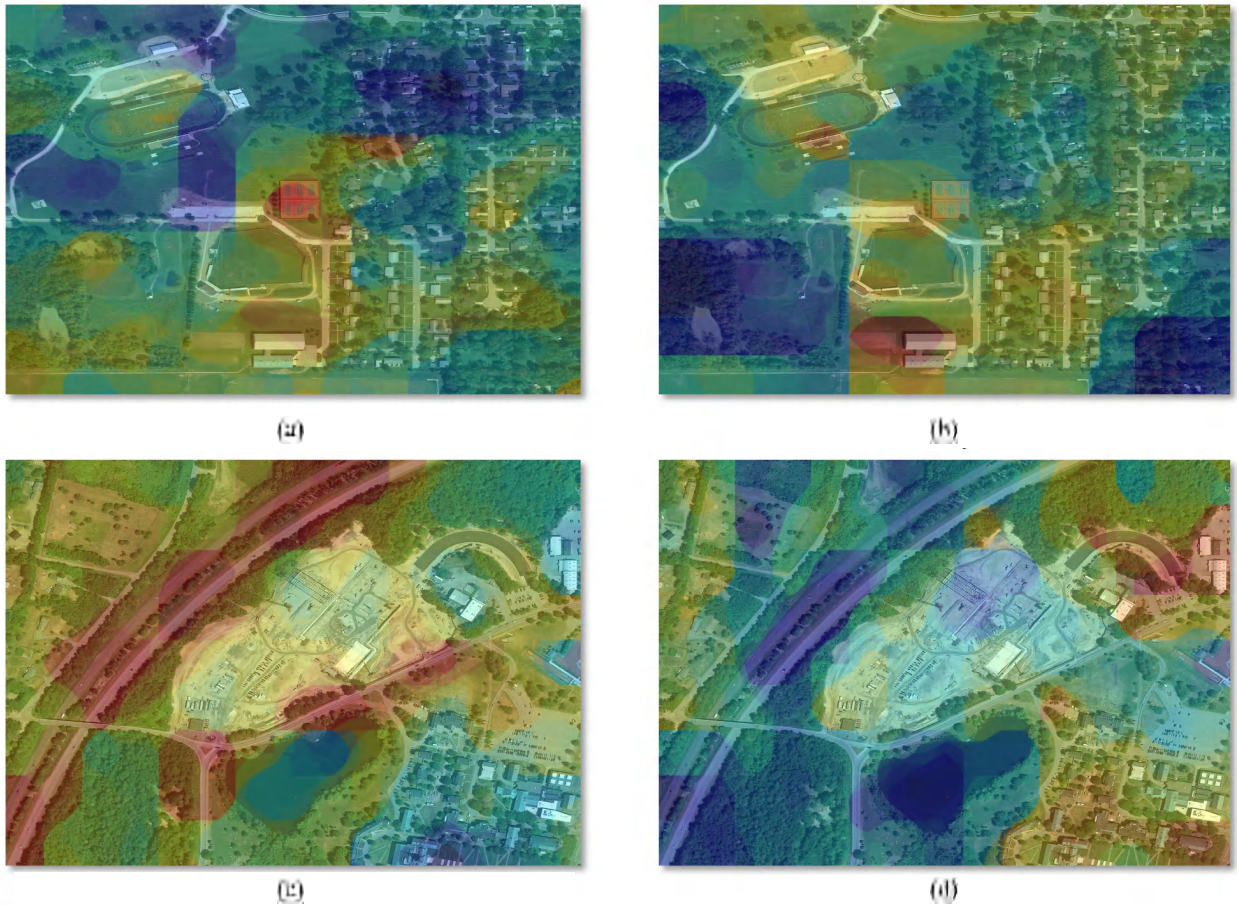

Fig. 14. Exploration results of location by text. (a) Query: there is a row of trees on one side of the six tennis courts and a white road beside it. (b) Query: a brown building rectangular beside a white spire. (c) Query: there is a highway next to the yellow bare ground with trees planted on both sides of highway. (d) Query: there are some vehicles parked on one side of the curved gray rod.

slight attention, the model puts all the remaining attention on the ground truth.

The above experiment shows that AMFMN can locate RS images effectively. AMFMN can adapt to more kinds of inputs and can also detect the adjacency between multiple objects even if our results are not as accurate as specific subtasks, such as target detection and road extraction. Therefore, compared with some specific retrieval tasks, RS text–image retrieval is a higher level retrieval task. Even though the accuracy of this task has not yet reached the ideal level by now, with the development of technology, this method will be inevitably applicable with vitality.

### I. Analysis of Time Consumption

In this section, we compare the evaluation time and the inference time of AMFMN on the four datasets to provide a baseline for subsequent research. Evaluation time refers to the total time spent in testing on a specific dataset. Due to the inconsistent size of the dataset, the evaluation time is also different. The inference time refers to the time spent of the model in one cross-modal information similarity calculation. For every results, we run several experiments to make the recorded results as accurate as possible.

TABLE VII
TIME-CONSUMING COMPARISON OF DIFFERENT METHODS

| Approach | Evaluation time(s) | | | | Inference time(s) |
|---|---|---|---|---|---|
| | RSICD | RSITMD | UCM | Sydney | |
| VSE++ | 52.59 | 10.54 | 4.01 | 1.67 | 0.0058 |
| SCAN | 76.55 | 14.46 | 5.51 | 1.98 | 0.1025 |
| MTFN | 69.83 | 12.07 | 4.20 | 1.85 | 0.0722 |
| CAMP | 54.01 | 11.63 | 4.63 | 1.88 | 0.0610 |
| AMFMN | 58.57 | 12.14 | 4.22 | 1.81 | 0.0681 |

Table VII shows the time-consuming comparison of different methods. Although VSE++ has a definite advantage in terms of model simplicity, at the same time, the minimalist VSE++ also affects the performance of the model to some extent. For MTFN, CAMP, and the proposed AMFMN method, the corresponding inference time is almost at the same level. The SCAN method consumes the highest time, which reflects that the calculation of this method may be more complicated. The response of the proposed method responds in seconds when performing retrieval in the database, belonging to the ordinary real-time system [8].

## VI. Conclusion

This article proposes an RS retrieval framework, which directly calculates the similarity between the RS images and the query text. We mainly establish an asymmetric multi-modal feature matching model suitable for multisource input and contribute a fine-grained RSITMD. The results show that for RS images with multiscale and redundant targets, the retrieval accuracy can be improved by using MVSA to extract the visual features and using VGMF to guide the text representation. The qualitative and quantitative results on several RS image-text datasets indicate that direct calculation of image and text similarity may be the next frontier for RS retrieval.

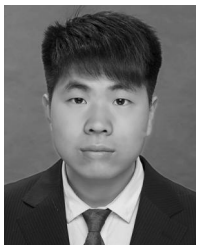

**Zhiqiang Yuan** (Student Member, IEEE) received the B.Sc. degree from Harbin Engineering University, Harbin, China, in 2019. He is pursuing the Ph.D. degree with the Aerospace Information Research Institute, Chinese Academy of Sciences, Beijing, China.

His research interests include computer vision, pattern recognition, and remote sensing image processing, especially on cross-modal retrieval and image caption.

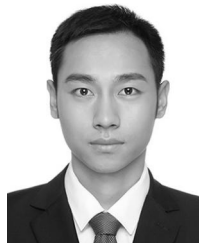

**Wenkai Zhang** (Member, IEEE) received the B.Sc. degree from the China University of Petroleum, Qingdao, China, in 2013, and the Ph.D. degree from the Institute of Electronics, Chinese Academy of Sciences, Beijing, China, in 2018.

He is an Assistant Professor with the Aerospace Information Research Institute, Chinese Academy of Sciences. His research interests include remote sensing image semantic segmentation and multimodel information processing.

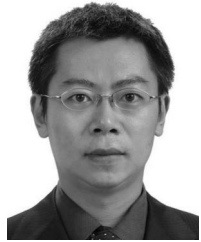

**Kun Fu** (Member, IEEE) received the B.Sc., M.Sc., and Ph.D. degrees from the National University of Defense Technology, Changsha, China, in 1995, 1999, and 2002, respectively.

He is a Professor with the Aerospace Information Research Institute, Chinese Academy of Sciences, Beijing, China. His research interests include computer vision, remote sensing image understanding, geospatial data mining, and visualization.

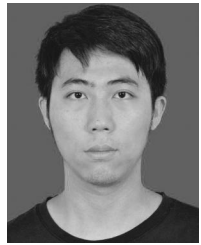

**Xuan Li** received the B.Sc. degree from Jilin University, Changchun, Jilin, China, in 2017. He is pursuing the Ph.D. degree with the Aerospace Information Research Institute, Chinese Academy of Sciences, Beijing, China.

His research interests include computer vision, pattern recognition, and multimodal learning, especially on image caption.

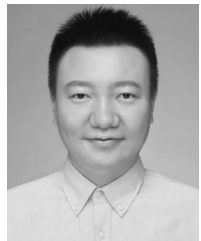

**Chubo Deng** received the B.Sc. degree from Hong Kong Baptist University, Hong Kong, in 2012, and the M.Sc. and Ph.D. degrees from George Washington University, Washington, DC, USA, in 2018.

He is an Assistant Professor with the Aerospace Information Research Institute, Chinese Academy of Sciences, Beijing, China. His research interests include computer vision, geospatial data mining, and remote sensing image understanding.

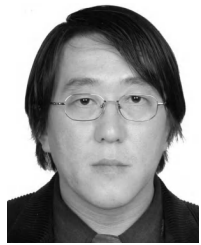

**Hongqi Wang** (Member, IEEE) received the B.Sc. degree from the Changchun University of Science and Technology, Changchun, China, in 1983, the M.Sc. degree from the Changchun Institute of Optics, Fine Mechanics and Physics, Chinese Academy of Sciences, Changchun, in 1988, and the Ph.D. degree from the Institute of Electronics, Chinese Academy of Sciences, Beijing, China, in 1994.

He is a Professor with the Aerospace Information Research Institute, Chinese Academy of Sciences. His research interests include computer vision and remote sensing image understanding.

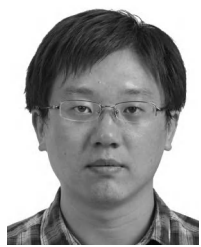

**Xian Sun** (Senior Member, IEEE) received the B.Sc. degree from the Beijing University of Aeronautics and Astronautics, Beijing, China, in 2004, and the M.Sc. and Ph.D. degrees from the Institute of Electronics, Chinese Academy of Sciences, Beijing, in 2009.

He is a Professor with the Aerospace Information Research Institute, Chinese Academy of Sciences. His research interests include computer vision, geospatial data mining, and remote sensing image understanding.